\documentclass[nohyperref]{article}

\usepackage{microtype}
\usepackage{graphicx}
\usepackage{subfigure}
\usepackage{booktabs} 
\usepackage{float}

\usepackage{natbib}

\usepackage{hyperref}

\usepackage[accepted]{icml2022}

\usepackage{amsmath}
\usepackage{amssymb}
\usepackage{mathtools}
\usepackage{amsthm}

\usepackage[capitalize,noabbrev]{cleveref}

\theoremstyle{plain}

\theoremstyle{definition}

\theoremstyle{remark}

\usepackage[textsize=tiny]{todonotes}

\icmltitlerunning{Riemannian data-dependent randomized smoothing for neural networks certification}

\begin{document}

\twocolumn[
\icmltitle{Riemannian data-dependent randomized smoothing for neural networks certification}



\icmlsetsymbol{equal}{*}

\begin{icmlauthorlist}
\icmlauthor{Pol Labarbarie}{yyy}
\icmlauthor{Hatem Hajri}{yyy}
\icmlauthor{Marc Arnaudon}{comp}

\end{icmlauthorlist}

\icmlaffiliation{yyy}{Institut de Recherche Technologique SystemX, Palaiseau, France}
\icmlaffiliation{comp}{Univ. Bordeaux, CNRS, Bordeaux INP, IMB, UMR 5251, Talence, France}

\icmlcorrespondingauthor{Pol Labarbarie}{pol.labarbarie@irt-systemx.fr}

\icmlkeywords{Machine Learning, ICML}

\vskip 0.3in
]



\printAffiliationsAndNotice{} 

\begin{abstract}
Certification of neural networks is an important and challenging problem that has been attracting the attention of the machine learning community since few years. In this paper, we focus on randomized smoothing (RS) which is considered as the state-of-the-art method to obtain certifiably robust neural networks. In particular, a new data-dependent RS technique called ANCER introduced recently can be used to certify ellipses with orthogonal axis near each input data of the neural network. In this work, we remark that ANCER is not invariant under rotation of input data and propose a new rotationally-invariant formulation of it which can certify ellipses without constraints on their axis. Our approach called Riemannian Data Dependant Randomized Smoothing (RDDRS) relies on information geometry techniques on the manifold of covariance matrices and can certify bigger regions than ANCER based on our experiments on the MNIST dataset. Code and models are available at \url{https://github.com/PLabarbarie/RDDRS}. 
\end{abstract}

\section{Introduction}
Making neural networks robust against adversarial examples \cite{szegedy2013intriguing,biggio2013evasion} or proving that they are has been the focus of many works. Robustification methods can be split into two main categories. First, empirical defenses empirically improve the resilience of neural networks but has no theoretical guarantee. The most and only successful known empirical defense technique is adversarial training \cite{madry2017towards}. Second, certified defense techniques try to provide a guarantee that an adversary does not exist in a certain neighborhood around a given input  \cite{huang2017safety,raghunathan2018semidefinite,bunel2018unified,lomuscio2017approach,tjeng2017evaluating,weng2018towards, lecuyer2019certified, cohen2019certified}. Despite the progress on these methods, the certified regions are still meaningless compared with the human perception.

In this paper we focus on the Randomized smoothing (RS) technique \cite{lecuyer2019certified, cohen2019certified} which, despite its simplicity, has proved the state-of-the-art to certify neural networks on small neighborhoods of inputs. Given a base classifier $F$, RS constructs a new classifier $G = F * \mathcal{N}(0,\sigma^2I_d)$. The smoothed classifier is certified robust: $G(x) = G(x+\delta)$ for all $\delta$ such as $||\delta||_2 < R(x,\sigma)$ with 

\begin{equation}\label{tt}
    R(x,\sigma)=\frac{\sigma^2}{2}\left(\Phi^{-1}(p_A) - \Phi^{-1}(p_B)\right)
\end{equation}

where $p_A = \mathbb{E}_Z[F_{c_A}(x+Z)] = G(x)_{c_A}$ is the probability of the top class $c_A$, $p_B = \max_{c \neq c_A} \mathbb{E}_Z[F_{c}(x+Z)] = G(x)_{c_B}$ is the probability of the runner-up class $c_B$, $Z \sim \mathcal{N}(0,\sigma^2I_d)$ and $\Phi^{-1}$ is the inverse cumulative distribution function of a standard Gaussian distribution. Several extensions and improvements of RS have been proposed subsequently. For instance, by adversarially training the base classifier \cite{salman2019provably}, regularization \cite{zhai2020macer}, general adjustments for training routines \cite{zhai2020macer} and by extending RS to more general distributions beyong the Gaussian case \cite{yang2020randomized}.

One limitation of the original RS work \cite{cohen2019certified} is that it only certifies isotropic regions. Properly extending this approach to anisotropic domains is a first important challenge. Enlarging at most the certified regions is a second ongoing challenge for all the proposed approaches. To generalise RS beyond fixed variance noise, \cite{alfarra2020data} 
proposed data-dependent randomised smoothing (DDRS) which additionally maximises the radius (\ref{tt}) over the parameter $\sigma$. This idea has been improved in \cite{eiras2021ancer} which proposes ANCER; an anisotropic diagonal certification method that performs sample-wise ($i.e.$ per sample in the test set) region volume maximization. They generalized (\ref{tt}) for ellipsoid regions $i.e$ they proved that for $\Sigma$ a non degenerate covariance matrix, $G(x) = G(x+\delta)$ for all $\delta$ such as $||\delta||_{\Sigma,2} < r(x,\Sigma)$ with 

\begin{equation}\label{tt2}
    r(x,\Sigma)=\Phi^{-1}(p_A) - \Phi^{-1}(p_B)
\end{equation}

and where $G = F * \mathcal{N}(0,\Sigma)$. 

\begin{figure*}[ht]
\includegraphics[width=1\textwidth]{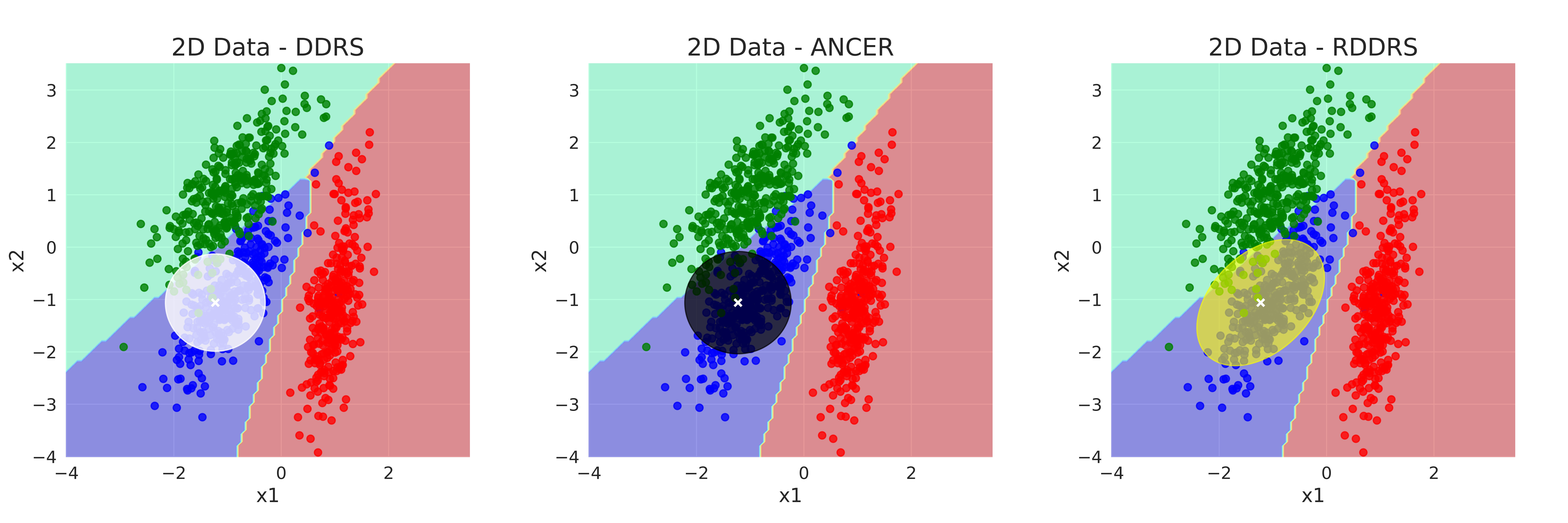}
\vskip -0.2in
\caption{Representation of a 2D classification problem in 3 classes (red, blue, and green). For the same point (white cross) from the test set, we run DDRS, ANCER and RDDRS. We plot the boundaries of the isotropic sampling region obtained with DDRS (white), anisotropic diagonal sampling region obtained with ANCER (black), and general anisotropic sampling region obtained with RDDRS (yellow).}
\label{2D DATA}
\end{figure*}

\noindent\textbf{Motivation and contribution.} The main motivation of our work is that ANCER is not resilient to rotations of input data. Indeed, in a configuration where the input data are concentrated around one canonical axis, ANCER may give good results. However, rotating all the points will dramatically impact the method since it always seeks for axis parallel to the canonical ones which will not be optimal after rotation. We propose an approach based on the use of information geometry which adequately fixes this issue by design as depicted by Figure \ref{2D DATA}. To summarize our approach, we consider a general Gaussian noise $\mathcal N(0,C)$ with covariance $C$ and for each input $x$ compute the corresponding certification radius $R(x,C)$. Next, we maximize $R(x,C)$ over $C$ on the manifold of covariance matrices through Riemannian optimisation and finally smooth the original classifier with the noise $\mathcal N(0,C)$. This methodology generalises ANCER which assumes that $C$ is a diagonal matrix and optimizes its diagonal in the Euclidean space. Through experiments on MNIST \cite{lecun1998gradient}, we show that our algorithm outperforms the previous works \cite{alfarra2020data,eiras2021ancer} and obtain new state-of-the-art certified accuracy for the DDRS technique.






\noindent\textbf{Organisation of the paper.} In Section \ref{rt}, we review existing DDRS techniques. In Section \ref{our}, we present our approach named Riemannian DDRS. In Section \ref{ex}, we experiment our method and show its advantage with respect to the state-of-the-art. Finally, in Section \ref{conc}, we discuss limitations of our method and RS techniques in general and outline some challenges related to them. 

\textbf{Notations.} We denote by $F : \mathbb{R}^d \longrightarrow \mathcal{P}(\mathcal{Y})$ a base classifier typically a neural network, where $\mathcal{P}(\mathcal{Y})$ is a probability simplex over $K$ classes. $\Phi^{-1}$ is the inverse cumulative distribution function of a standard Gaussian distribution. The $\ell_p$-ball is defined with respect to the $||.||_p$ norm ($p \geq 1$) and the $\ell_p^A$-ellipsoid is defined with respect to $||.||_{A,p}$. For $p \in \{1,2\}$, $||.||_{A,p}$ is the composite norm defined with respect to a positive definite matrix $A$ and a vector $u$ as $||A^{-1/p}u||_p$.
\section{Data-dependent randomized smoothing}\label{rt}
DDRS \cite{alfarra2020data} aims to improve the choice of the hyperparameter $\sigma$ in order to obtain bigger certified balls. This parameter is supposed to be fixed by the original RS method. Concretely, DDRS solves the optimisation problem 
\begin{equation}\label{DDS_equation}
\sigma_x^* = \arg\max_\sigma R(x,\sigma)
\end{equation}
using gradient ascent and defines the smoothed classifier $G$ by means of the sampling noise $\mathcal N(0,{\sigma_x^*}^2 I)$ for each input $x$. This approach showed better certified isotropic regions per points. Next, \cite{eiras2021ancer} generalized randomized smoothing to certify anisotropic regions, by pairing it with a generalization of the data-dependent method to maximize the certified region at each input point. Following a Lipschitz argument of \cite{salman2019provably,jordan2020exactly}, they developed a methodology to compare anisotropic regions with isotropic ones. They proposed the following optimisation problem
\begin{equation}\label{ANCER_equation}
\arg\max_{\Theta^x}  r(x,\Theta^x) \sqrt[d]{\prod_i \theta_i^x}
\end{equation}
under the constraint that $\min_i \theta_i^x r(x,\Theta^x) \geq r_{iso}^*$ where $r(x,\Theta^x) = \Phi^{-1}(p_A) - \Phi^{-1}(p_B)$, $r_{iso}^*$ is the maximum isotropic $l_2$ radius obtained with (\ref{DDS_equation}) and $\Theta_x  = \text{diag}(\{ \theta_i\}_{i=1}^d)$. The constraint ensures that the isotropic region obtained by (\ref{DDS_equation}) is enclosed in the anisotropic one.

\section{Riemannian data-dependent randomized smoothing}\label{our}
As outlined previously, we will consider general smoothing noises $\mathcal N(0,C)$ where $C$ is a covariance matrix, i.e a symmetric positive definite one. For this, we will work on the space $\mathcal{S}_d^{++}$ of $d\times d$ covariance matrices which will be equipped with the Riemannian metric known as the affine-invariant metric or Rao–Fisher metric in information geometry \cite{10.2307/25050283,Skovgaard1984ARG}:
\begin{equation*}
    g_Y(A,B) = \text{tr}(Y^{-1}AY^{-1}B)
\end{equation*}
where $Y \in \mathcal{S}_d^{++}$ and $A,B \in T_Y\mathcal{S}_d^{++}$, with $T_Y\mathcal{S}_d^{++}$ is the tangent space to $\mathcal{S}_d^{++}$ at $Y$, identified with the space of symmetric $d \times d$ matrices. The Riemannian distance associated with this metric is  

\begin{equation*}
d^2(Y,Z) = \text{tr}(\log(Y^{-\frac{1}{2}}ZY^{-\frac{1}{2}}))^2
\end{equation*}

for all $Y, Z\in \mathcal{S}_d^{++}$. Next, we consider the following optimisation problem on $\mathcal{S}_d^{++}$:
\begin{equation*}
\begin{aligned}
& \arg\max_{C^x}
& & R(C^x) \sqrt[d]{\prod_i \lambda_i^x} \\
& \text{such that}
& & \min_i \lambda_i^x R(C^x) \geq r_{iso}^*\ \text{and}\ \lambda_i^x \geq \sigma_x^*
\end{aligned}
\end{equation*}
and where 

\begin{equation*}
R(C^x) = \left( \Phi^{-1}\left( G(C^x)_{c_A} \right) - \Phi^{-1}\left( G(C^x)_{c_B} \right) \right),
\end{equation*}

$G(C) = \mathbb{E}[F(x + CZ)]$ with $Z \sim \mathcal{N}(0,I_d)$ and the $\lambda_i^x$ are the eigenvalues of the matrix $C^x$. By using a penalisation method, we instead consider the maximisation problem:
\begin{equation}\label{pen}
    \arg\max_{C^x} \left(R(C^x) P(C^x) + \kappa K(C^x) R(C^x) \right)
\end{equation}

such that for all $i \in \{1,...,d\}$, $\lambda_i \geq \sigma_x^*$ and where: 

$$P(C^x) = \sqrt[d]{\prod_i \lambda_i^x},\  K(C^x) = \min_i \lambda_i^x$$

Define $H(C^x) = R(C^x) P(C^x) + \kappa K(C^x) R(C^x)$. Then, we can use Riemannian gradient descent to locally maximise $H$ on $\mathcal{S}_d^{++}$ by iterating 
\begin{equation}\label{update}
C_{t+1}^x = \text{Exp}_{C_t^x}(\gamma_t \nabla H(C_t^x))
\end{equation}

where $\gamma_t$ is a step size depending on $t$ and $\text{Exp}$ is the exponential map:  

\begin{equation*}
\text{Exp}_Y(\Delta) = Y^{\frac{1}{2}}\exp(Y^{-\frac{1}{2}}\Delta Y^{-\frac{1}{2}} )Y^{\frac{1}{2}}
\end{equation*}

The update (\ref{update}) requires computing $\nabla H$ which is given by:

$$\nabla H(C_t^x) = \nabla R(C_t^x)P(C_t^x) + R(C_t^x)\nabla P(C_t^x) $$
$$+ \kappa\left( \nabla R(C_t^x)K(C_t^x) + R(C_t^x)\nabla K(C_t^x) \right)  $$

with 

$$\nabla R(C_t^x) = \left( \frac{\nabla (G(C^x_t)_{c_A})}{\Phi'(\Phi^{-1}(G(C^x_t)_{c_A}))} - \frac{\nabla (G(C^x_t)_{c_B})}{\Phi'(\Phi^{-1}(G(C^x_t)_{c_B}))}   \right) $$

Let us now give the Riemannian gradient of $G$. If $Z \sim \mathcal{N}(0,I_d)$, we have $\nabla G(C) = C\mathbb{E}[A]C$, where: 

$$A_{ij} = \frac{1}{2} (Z_j(\nabla F(y))_i + Z_i(\nabla F(y))_j) $$
and $y := x + CZ$. Using this expression, we can approximate $\mathbb{E}[A]$ by Monte-Carlo and then deduce $\nabla G(C)$. Similarly, $\nabla P(C) = CBC$ and $\nabla K(C) = CDC$ where 
$$B_{ij} = \frac{1}{2} ((\nabla P(C))_i + (\nabla P(C))_j) $$

and

$$D_{ij} = \frac{1}{2} ((\nabla K(C))_i + (\nabla K(C))_j) $$

\begin{figure*}[!ht]
\includegraphics[width=1\textwidth]{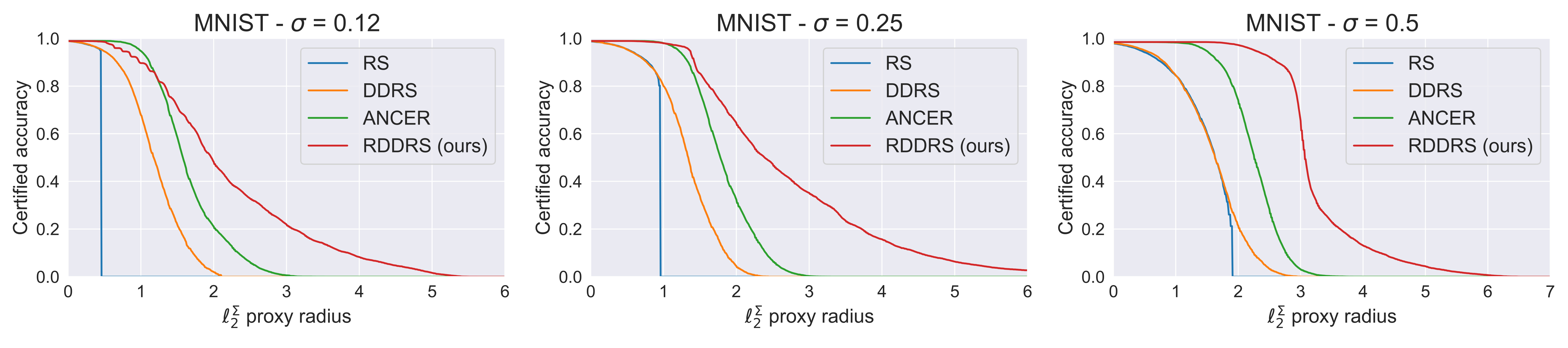}
\vskip -0.2in
\caption{Certified accuracy as a function of the proxy radius $\ell_{2}^\Sigma$ for several $\sigma$ used in training for MNIST dataset. We compare our Riemannian framework RDDRS, $\Sigma = C^x$, against the standard randomized smoothing (RS), $\Sigma = \sigma^2I_d$, the data-dependent randomized smoothing (DDRS), $\Sigma = \sigma_x^2I_d$,  and the anisotropic diagonal data-dependent randomized smoothing (ANCER), $\Sigma = \Theta^x$.}
\label{Certified accuracy}
\end{figure*}

\section{Experiments}\label{ex}
In this section, we empirically test the performance of our RDDRS framework for neural networks certification and show that it achieves state-of-the-art performance in terms of certified regions on the well-known MNIST dataset. The methods used for comparison are randomized smoothing (RS) as it appeared in \cite{cohen2019certified}, DDRS \cite{alfarra2020data} and ANCER \cite{eiras2021ancer}. For evaluation, the same procedure used in \cite{cohen2019certified,salman2019provably,alfarra2020data,eiras2021ancer} is followed. We train a classical convolutional neural network on MNIST using the Gaussian data augmentation proposed by \cite{cohen2019certified} with $\sigma$ varying in $\{0.12,0.25,0.5\}$ and then perform the certification on the whole MNIST test set. To compare between the different tested methods, we plot the approximate certified accuracy curve similar to previous works. This curve is computed for many radius $R$ and corresponds for each $R$ to the certified accuracy given by the fraction of the test set images which $G$ classifies correctly and certifies robust with a radius $R'\geq R$. During certification by the state-of-the-art methods, we use the same $\sigma$ as for training. To run these methods, we closely follow the recommendations of the respective papers. The DDRS optimization of (\ref{DDS_equation}) is done using a gradient ascent, a learning rate $\alpha = 0.0001$, an initial $\sigma_0 = \sigma$ and a number of samples $n=100$ for Monte-Carlo estimation of $p_A$ at each iteration. Note that following the previous works there is no need to estimate $p_B$ even if the number of classes is greater than $2$ (see \cite{cohen2019certified}). The number of gradient ascent iterations for DDRS is taken from $K=100$ to $K=1500$ with a step of $100$ and the $\sigma$ providing the best certified radius is saved. For ANCER, we resolved (\ref{ANCER_equation}) with different values of the learning rate $\alpha \in \{0,04,0,4\}$, a penalization parameter $\kappa = 2$, the same number of iterations $K=100$ and estimate $p_A$ with $n=100$. Our method RDDRS is run with different values of the learning rate $\gamma_t \in \{0,5,1,25\}, \kappa = 10^{-6}$, the same number of iterations $K=100$ and a number of samples $n=20000$ to estimate $\mathbb{E}[A]$. To calculate the exponential map on $\mathcal{S}_d^{++}$, we use the Geomstats library \cite{JMLR:v21:19-027, miolane:hal-02908006}. After obtaining a first trained model for three valued of $\sigma$ as previously described, we run the four methods RS, DDRS, ANCER and RDDRS and report the final certified accuracy. The obtained results are plotted in Figure \ref{Certified accuracy} and commented in what follows.

\textbf{Comments.} The graphics of the certified accuracy obtained for the four methods and the three values of $\sigma$ demonstrate that our RDDRS method globally outperforms DDRS and ANCER for almost all proxy radius and for all the values of $\sigma$. The difference becomes more important in our favour when $R$ gets bigger, $i.e$ $R>3$ and also when $\sigma$ increases. Only for $\sigma=0.12$ and small radius $R$ our method is slighlty less better than ANCER. Moreover, we observe that none of the certification methods except ours can certify a good percentage of the test set at large radii. All of this illustrates that our method has significantly pushed forward the data-dependent smoothing techniques.

\textbf{Discussion on the parameter $\kappa$.} \cite{eiras2021ancer} proposed to penalize (\ref{ANCER_equation}) for empirical considerations. They set the default value of $\kappa$ to 2. For our formulation (\ref{pen}), we have conducted several runs of RDDRS with different values of $\kappa$ and have concluded that this factor does not have a significant effect on the method's convergence. The lower bound condition $\lambda_i^x \geq \sigma_x^*$ has however a more  significant effect on that convergence.

\textbf{Runtime discussion. } Our method is slower than the other methods. This essentially comes from the computation of $\mathbb E[A]$ for which we considered a greater number of samples $n=20000$. Indeed, despite the fact that this value is relatively high, it allows our results to be very stable: different runs with the same $n$, give almost the same performances. 

\textbf{Scalability challenges.} Computing the exponential map becomes costly for high dimensional matrices. We believe that, on datasets such as CIFAR10 \cite{cifar}, these computations are feasible and our method is still applicable with reasonable computation ressources. However, extending the approach to high-dimensional data such as IMAGENET \cite{5206848} seems to be very challenging. Some geometric approximations of the exponential map could be useful for that.  

\section{Limitations and Conclusion}\label{conc}
Randomized smoothing is currently the state-of-the-art method to certify neural networks despite its curse of dimensionality limitation \cite{osti_10181769}. In this paper, we improved the data-dependent RS technique by looking for the best Gaussian covariance matrix to define the smoothed classifier. Our method is less scalable than other approaches but performs significantly better on MNIST which is an important dataset in deep learning. We set up several challenges in the frontier between differential geometry and deep learning to push forward this approach to high-dimensional data. We relied on the Rao-Fisher distance, but other metrics could have been used such as the log-Euclidean metric on the space of covariance matrices \cite{Arsigny2006LogEuclideanMF}. It seems to us an important question to test our method in terms of cost and also certification regions as a function of the used distance. Regarding complexity, it should be remarked that RS networks are in general slower than classical convolutional neural network. Data dependent smoothing networks are even slower. Also, they are little bit ambiguous; indeed, one network is defined and optimised for each input data. To adversarially attack the smoothed network at a given input, one has to attack the smoothed network at this point; defined with the optimal noise found in this paper. While this procedure is much more computational, it is more robust as it guarantees that adversarial examples can not be found in the certified regions. Finally, extending data dependent approaches beyond the $L_2$ metric such as to defend against $L_0$ attacks (see for instance the works \cite{Papernot2016TheLO,make2040030,9643161}) seems to us an important challenge. 

\section*{Acknowledgements}
This work has been supported by the French government under the France 2030 program, as part of the SystemX Technological Research Institute. 


\nocite{langley00}

\bibliography{example_paper}
\bibliographystyle{icml2022}


\end{document}